\documentclass[letterpaper]{article} 

\usepackage{aaai2027}
\usepackage[hyphens]{url}  
\usepackage{graphicx} 
\usepackage{natbib}  
\usepackage{caption} 
\usepackage{algorithm}
\usepackage{algorithmic}
\usepackage{amsmath}

\usepackage{newfloat}
\usepackage{listings}
\DeclareCaptionStyle{ruled}{labelfont=normalfont,labelsep=colon,strut=off} 
\floatstyle{ruled}
\newfloat{listing}{tb}{lst}{}
\floatname{listing}{Listing}

\usepackage{booktabs}

\usepackage[utf8]{inputenc} 
\usepackage[T1]{fontenc}    
\usepackage{url}            
\usepackage{booktabs}       
\usepackage{amsfonts}       
\usepackage{nicefrac}       
\usepackage{microtype}      
\usepackage{xcolor}         

\usepackage{graphicx}
\usepackage{svg}
\usepackage{makecell}
\usepackage{multirow}
\usepackage{amsmath}
\usepackage{caption}
\usepackage{colortbl}
\usepackage{tabularx}

\usepackage{tcolorbox}
\tcbuselibrary{skins, breakable}

\usepackage{booktabs}
\usepackage{multirow}
\usepackage{siunitx}

\title{SkyNative: A Native Multimodal Architecture for Remote Sensing Vision-Language Understanding}

\author{
    Xiao Yang\textsuperscript{\rm 1,2},
    Ronghao Fu\textsuperscript{\rm 1,2}\corresponding,
    Zhiwen Lin\textsuperscript{\rm 1,2},
    Zhuoran Duan\textsuperscript{\rm 1,2},
    Lang Sun\textsuperscript{\rm 1,2},
    Jiaqi Liu\textsuperscript{\rm 1,2},
    Jiashun Zhu\textsuperscript{\rm 1,2},
    Jiasen Hu\textsuperscript{\rm 1,2},
    Xu Na\textsuperscript{\rm 1,2},
    Bo Yang\textsuperscript{\rm 1,2}\corresponding
}

\affiliations{
    \textsuperscript{\rm 1}College of Computer Science and Technology, Jilin University, China\\
    \textsuperscript{\rm 2}Key Laboratory of Symbolic Computation and Knowledge Engineering of Ministry of Education\\
}

\begin{document}

\maketitle

\begin{abstract}
Remote sensing vision-language models (RS-VLMs) commonly employ a pretrained vision encoder and a projection module to map image features into the token space of a large language model. Although effective, this modular RS-VLMs separates visual representation from language reasoning, potentially limiting the direct involvement of fine-grained visual evidence in complex spatial inference. This challenge is particularly relevant to remote sensing imagery, which often covers broad geographic areas and contains multi-scale objects, dense target distributions, and intricate spatial layouts. In this paper, we propose SkyNative, the first study to explore a native multimodal architecture for remote sensing vision-language tasks. SkyNative converts remote sensing images into visual tokens through a lightweight patch embedding module and places them together with text tokens in a shared autoregressive sequence, allowing textual tokens to directly access the preceding visual context. To accommodate the heterogeneous characteristics of the two modalities, we further adopt a modality-aware decoupling mechanism that applies modality-specific projections, normalization, and feed-forward transformations while processing both modalities through shared causal self-attention. Extensive experiments demonstrate SkyNative’s strong capabilities in dense small-object perception, large-format contextual understanding, complex reasoning, and robustness, with scores of 68.93\%, 47.40\%, and 63.43\% on HRRSD, RSHR reasoning, and OmniEarth, respectively.  These results suggest that the native VLM architecture explored in SkyNative represents a promising approach to RS vision-language modeling.

\end{abstract}

\section{Introduction}
\label{sec:introduction}

Remote sensing vision-language models (RS-VLMs) have made substantial progress in visual question answering, scene classification, image captioning, and instruction-following tasks~\cite{li2024vision,hong2026foundation}, moving remote sensing image understanding from task-specific modeling toward a unified vision-language framework. Unlike general images, remote sensing images often cover large-scale ground scenes with densely distributed multi-scale objects and complex spatial layouts. As a result, remote sensing interpretation tasks often require fine-grained object recognition, region-level semantic understanding, and scene-level spatial reasoning.

\begin{figure}[t]
    \centering
    \includegraphics[width=\columnwidth]{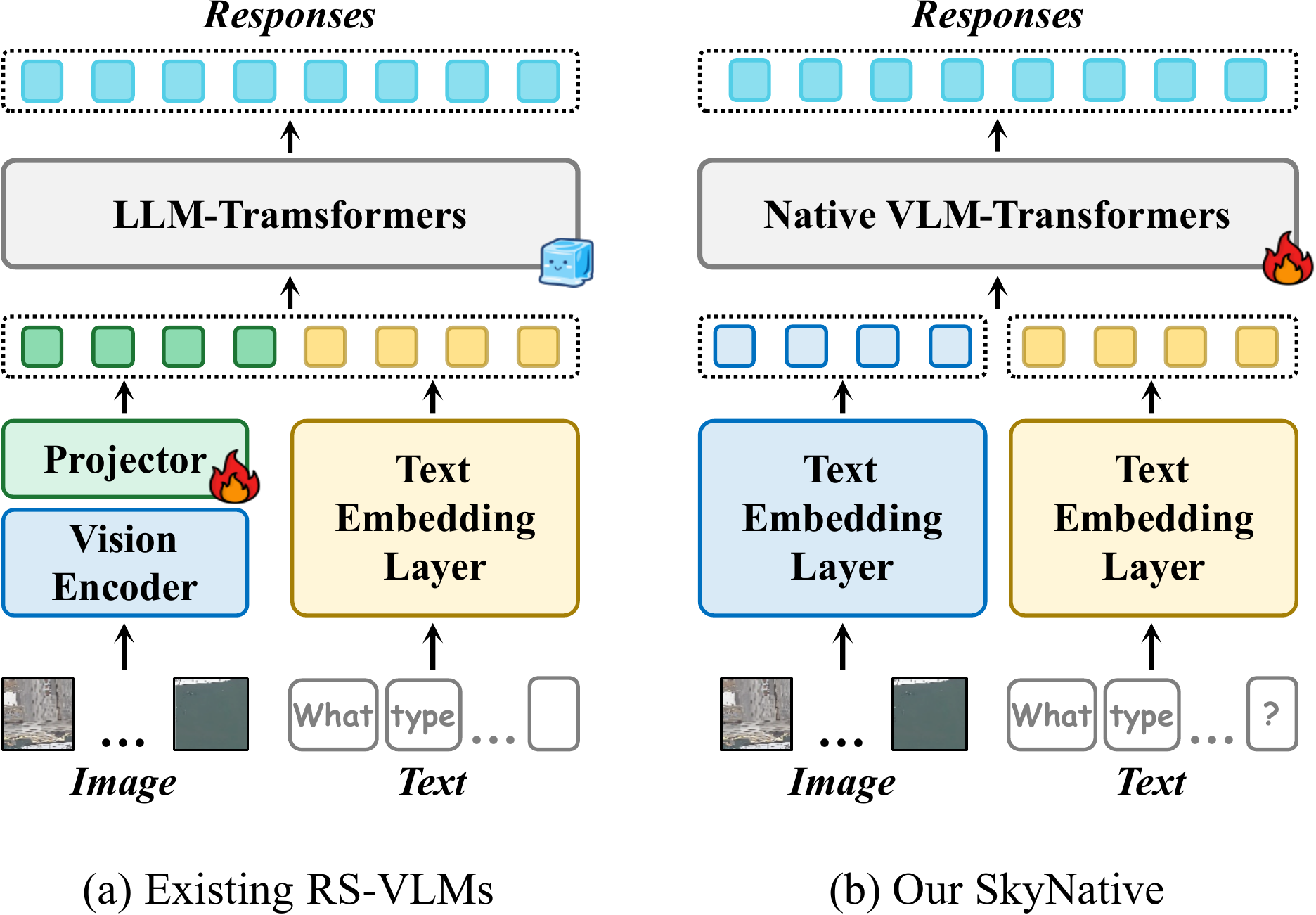}
    \caption{
    Comparison between the encoder-based RS-VLM architecture and our native multimodal SkyNative.
    }
    \label{fig:intro_arch}
\end{figure}

\begin{figure}[t]
    \centering
    \includegraphics[width=\columnwidth]{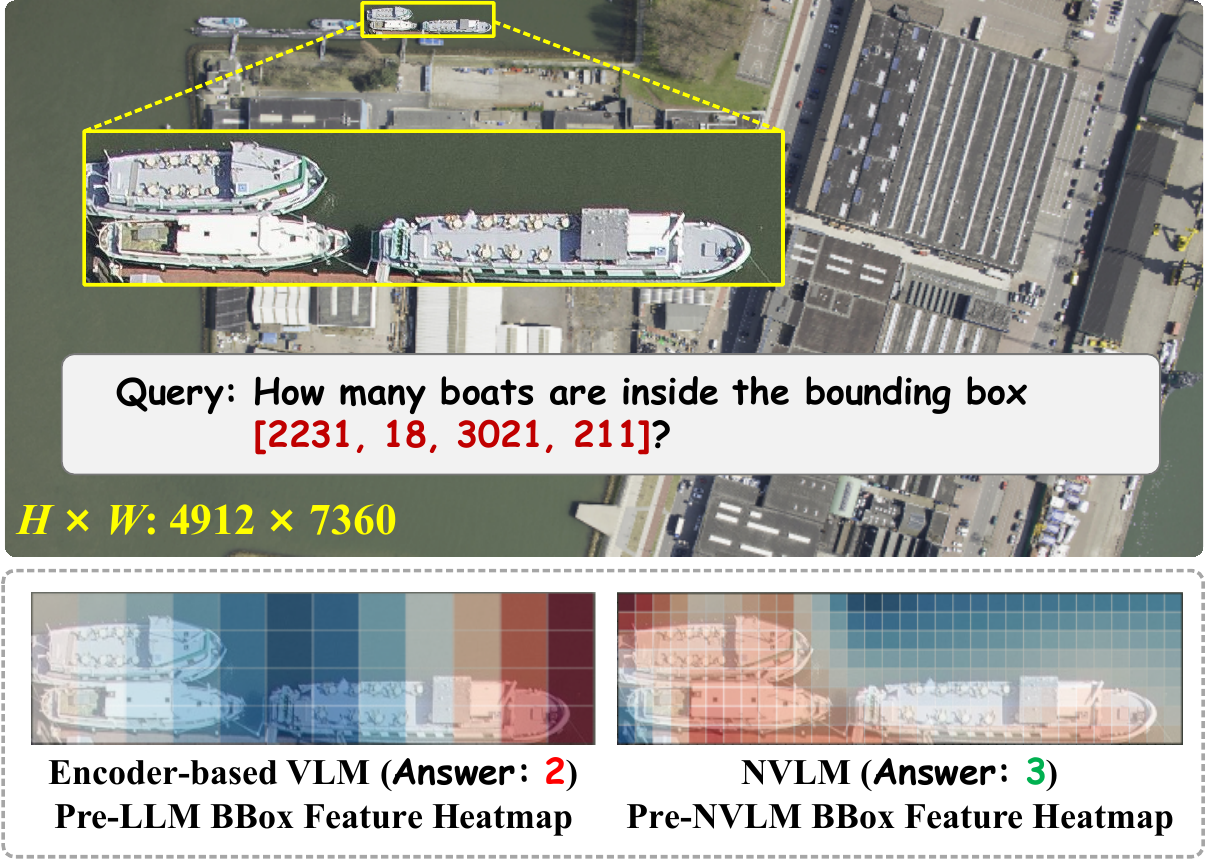}
    \caption{
    BBox-aligned feature heatmaps at the pre-LLM and pre-NVLM. The heatmaps visualize normalized feature responses within the query-specified bounding box. The encoder–projector architecture provides coarse pre-LLM spatial support, while the native architecture preserves a denser pre-NVLM representation.
    }
    \label{fig:intro_motivation}
\end{figure}

Existing RS-VLMs adopt an encoder-based architecture, as illustrated in Fig.~\ref{fig:intro_arch}(a). In this paradigm, a pretrained vision encoder extracts image features, and a projector maps them into the token space of a large language model (LLM) for cross-modal understanding and reasoning. This modular design is reusable and practical, making it a strong foundation for current RS-VLMs. However, the characteristics of remote sensing images pose additional requirements for this modeling interface. Small objects, dense land-cover patterns, and complex spatial layouts require visual representations with sufficient spatial granularity, as well as effective joint modeling between visual content and textual queries. As illustrated in Fig.~\ref{fig:intro_motivation}, region-level counting over large-format imagery relies on sufficiently localized spatial representations at the pre-LLM interface. This motivates us to explore a unified architecture in which visual representation, cross-modal interaction, and language reasoning are jointly modeled within the same token sequence.

Recent native multimodal models provide an alternative architectural perspective for vision-language modeling. Instead of connecting a separately pretrained vision encoder to a language backbone through post-hoc alignment, native multimodal models typically process image tokens and text tokens within a unified Transformer, allowing both modalities to be represented in a shared autoregressive token space and enabling textual tokens to access preceding visual context~\cite{an2026native}. The key to this paradigm is not merely to remove the separate vision encoder, but to provide a unified modeling interface for image and language through a shared token sequence, positional modeling, and causal self-attention. For RS tasks, this interface is well aligned with the need to make local objects, dense land-cover patterns, and spatial layouts directly accessible to text-conditioned reasoning.

Inspired by this paradigm, we make the first exploration of native multimodal architecture for remote sensing vision-language tasks and propose SkyNative. As shown in Fig.~\ref{fig:intro_arch}(b), SkyNative uses a lightweight image patch embedding layer to convert remote sensing images into native visual tokens, which are jointly processed with text tokens by a unified Autoregressive Transformer. To account for the different modeling characteristics of visual and textual tokens, SkyNative further employs a modality-aware decoupling mechanism that assigns separate parameter paths to the two modalities within the unified framework. We evaluate SkyNative on multiple remote sensing vision-language benchmarks, covering general remote sensing understanding, fine-grained object recognition, spatial relation analysis, and large-format image reasoning. Experimental results demonstrate the effectiveness of SkyNative for remote sensing vision-language modeling.

Our contributions are summarized as follows:

\begin{itemize}
    \item We make the first exploration of native multimodal architecture for remote sensing vision-language tasks and propose SkyNative.

    \item We adopt native visual tokenization and modality-aware decoupling to jointly model visual and textual tokens in a unified autoregressive Transformer.
    
    \item We evaluate SkyNative on multiple remote sensing vision-language benchmarks, demonstrating its effectiveness across diverse RS-VL tasks.
\end{itemize}

\section{Related Work}
\label{sec:related_work}

\paragraph{Remote Sensing Vision-Language Models.}
RS-VLMs have made significant progress in visual question answering, image captioning, scene classification, visual grounding, and instruction following. Representative works such as RSGPT~\cite{hu2025rsgpt}, GeoChat~\cite{kuckreja2024geochat}, and VHM~\cite{pang2025vhm} model the semantic associations between remote sensing images and natural language, and improve geospatial understanding through remote sensing image-text data and instruction tuning. Since remote sensing interpretation often involves fine-grained object recognition, large-scale scene understanding, and spatial relation reasoning, existing RS-VLMs are commonly evaluated on benchmarks for visual question answering, grounding, captioning, and high-resolution image understanding, which examine their ability to model multi-scale objects, dense scenes, and complex layouts.

Driven by these task requirements, recent studies further improve RS-VLMs from the perspectives of input formulation and region modeling. GeoChat~\cite{kuckreja2024geochat} processes remote sensing images through global rescaling, while GeoPixel~\cite{shabbir2025geopixel} adopts tile-and-concatenate operations to enhance local region coverage. GeoLLaVA-8K~\cite{wang2025geollava} and Coarse-to-fine~\cite{luo2025large} reduce the cost of long-sequence modeling through token pruning or staged perception, and ZoomEarth~\cite{liu2025zoomearth} dynamically searches query-relevant regions. These methods improve the handling of large-format inputs, dense land-cover patterns, and local objects from different perspectives, suggesting that RS-VLM architectures for the structural properties of remote sensing images remain worth further exploration.

\paragraph{Vision-Language Architectures.}
Vision-language modeling has progressed from contrastive image-text alignment to LLM-based multimodal reasoning. CLIP~\cite{radford2021learning} provides transferable visual representations and has become a common foundation for modern VLMs. Most existing methods, including BLIP~\cite{li2023blip} and LLaVA~\cite{liu2023visual}, follow an encoder--projector architecture: a pretrained vision encoder, such as CLIP or SigLIP~\cite{zhai2023sigmoid}, extracts image features, which are mapped into the LLM token space through an MLP projector or Q-Former~\cite{li2023blip}. This modular design has been widely adopted due to its strong performance and flexibility.

Native multimodal models offer an alternative by tokenizing visual inputs and processing visual and textual tokens as a common sequence within a shared Transformer backbone~\cite{an2026native}. Recent studies have explored continuous patch embeddings, encoder-free visual token modeling, and unified token prediction~\cite{li2026breen,wang2026emu3}. However, the suitability of native architectures for RS-VL tasks remains insufficiently studied. Such tasks often require fine-grained perception of small objects, spatial relationships, and densely distributed scene elements. Our work addresses this gap through native visual tokenization and modality-aware decoupling tailored to remote sensing imagery.

\begin{figure*}[t]
    \centering
    \includegraphics[width=\textwidth]{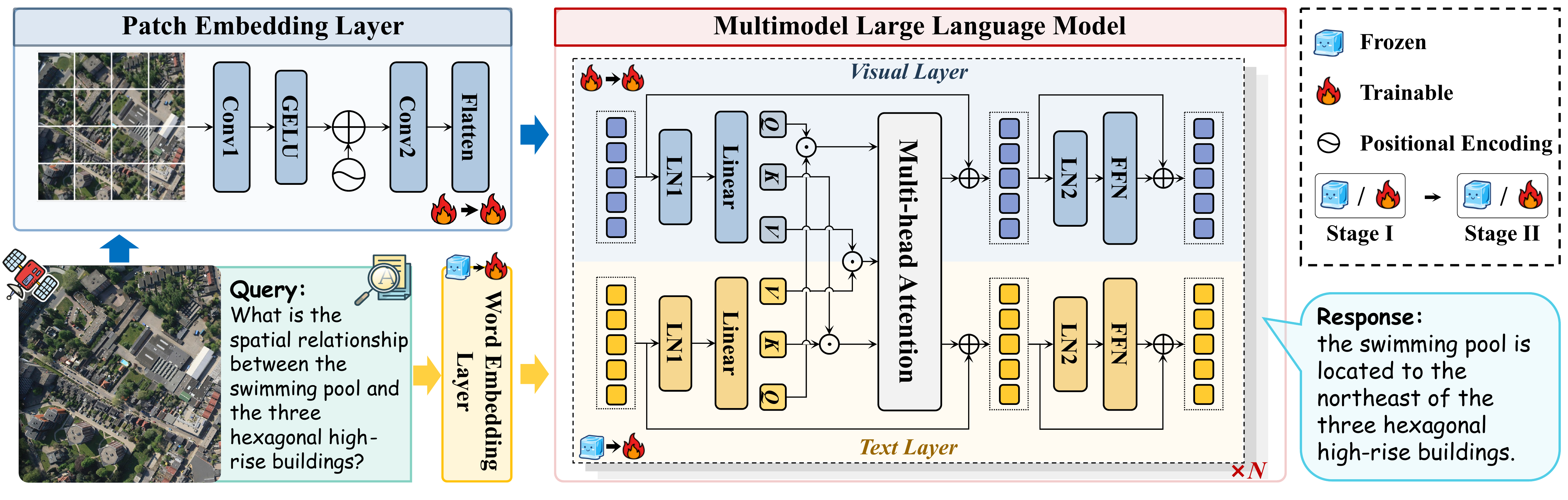}
    \caption{Overview of SkyNative framework.}
    \label{fig:architecture}
\end{figure*}

\section{Methodology}
\label{sec:methodology}

In this section, we present the overall architecture and training pipeline of the proposed SkyNative framework, as illustrated in Fig.~\ref{fig:architecture}. Unlike conventional RS-VLMs that rely on a standalone pretrained visual encoder, SkyNative follows a pretrained-vision-encoder-free paradigm. It maps native RS images into patch-level visual tokens through a shallow convolutional embedding layer, and feeds them together with textual token embeddings into a unified decoder-only Transformer for autoregressive reasoning.

\subsection{Vision-Language Encoding Layer}
\label{subsec:encoding}

To preserve fine-grained textures and spatial structures in RS imagery, SkyNative removes the standalone pretrained visual encoder and directly constructs visual tokens from native image patches. Given an input RS image $\mathbf{I}\in\mathbb{R}^{H\times W\times3}$, we employ a lightweight convolutional patch embedding layer consisting of two convolutional layers with a GELU activation in between:
\begin{equation}
\small
    z_v = \text{Conv2}(\text{GELU}(\text{Conv1}(\mathbf{I})))
    \in \mathbb{R}^{H' \times W' \times d},
\end{equation}
where $d$ denotes the hidden dimension of the language model. The resulting 2D feature map is then flattened into a sequence of visual tokens:
\begin{equation}
\small
    x_v = \text{Flatten}(z_v) + p_v,
\end{equation}
where $p_v$ denotes the visual positional embedding. This shallow mapping avoids compressing the image into high-level semantic features too early, thus preserving local visual evidence for subsequent reasoning.

For the text sequence $\mathbf{T}$, we use the tokenizer and embedding matrix of the base LLM:
\begin{equation}
\small
    x_t = E_t(\text{Tokenizer}(\mathbf{T})),
\end{equation}
where $E_t$ is the textual embedding matrix. The visual tokens $x_v$ are prepended to the textual tokens $x_t$ and processed as a single autoregressive sequence, allowing textual tokens to attend to the preceding visual context.

\subsection{Modality-Aware Decoupling Mechanism}
\label{subsec:decoupling}

Directly mixing low-level visual tokens and textual tokens in the same Transformer may cause cross-modal representation interference. To address this issue, we adopt a modality-aware decoupling mechanism that assigns modality-specific parameters to key Transformer components while retaining causal self-attention over the shared multimodal sequence.

Given a token sequence $x=(x_1,\ldots,x_n)$, each token $x_i$ is associated with a modality label $u_i\in\{v,t\}$, where $v$ and $t$ denote visual and textual modalities, respectively. For self-attention, the query, key, and value projections are computed with modality-specific weights:
\begin{equation}
\small
    Q_i=x_iW_Q^{u_i}, \quad
    K_i=x_iW_K^{u_i}, \quad
    V_i=x_iW_V^{u_i}.
\end{equation}

The projected tokens from both modalities are then processed within the same causal multi-head self-attention operation:
\begin{equation}
\small
    \text{MHA}(Q,K,V)
    =
    \text{softmax}
    \left(
    \frac{QK^{T}}{\sqrt{d_k}} + M
    \right)V,
\end{equation}
where $M$ denotes the causal attention mask. In this way, answer tokens can attend to visual evidence and previous textual tokens during autoregressive generation.

In addition to attention projections, we also use modality-specific layer normalization and FFN parameters. The Transformer block is formulated as:
\begin{equation}
\small
    h = x + \text{MHA}(\text{LN}_1^{u}(x); \theta_{\text{attn}}^{u}),
    \quad
    x' = h + \text{FFN}^{u}(\text{LN}_2^{u}(h)).
\end{equation}

This decoupled design allows visual and textual tokens to use modality-specific transformations while still interacting within a shared autoregressive backbone. During early training, the textual LLM parameters are kept frozen to preserve linguistic knowledge, while the newly introduced visual components learn to align native RS patches with the LLM token space.

\subsection{Training Objective}
\label{subsec:objective}

SkyNative is trained with the standard autoregressive next-token prediction objective. Given visual tokens $x_v$, instruction tokens $x_q$, and the target answer sequence $\mathbf{Y}=(y_1,\ldots,y_L)$, the model predicts each answer token conditioned on the visual context and previous textual tokens:
\begin{equation}
\small
    \mathcal{L}
    =
    -\sum_{i=1}^{L}
    \log p(y_i \mid x_v, x_q, y_{<i}; \Theta),
\end{equation}
where $\Theta$ denotes the trainable parameters. The loss is only computed on target answer tokens, while visual tokens and instruction tokens are masked out. This objective is consistently applied across all training stages.

\subsection{Training Strategy}
\label{subsec:training}

The training of SkyNative follows a two-stage progressive strategy, moving from visual token learning to geospatial instruction tuning.
\textbf{Stage I: Modality-Aware Alignment.}
We jointly train the patch embedding layer and the modality-specific visual parameters, while keeping the core textual parameters frozen. This stage aligns visual tokens with textual semantics and reduces cross-modal representation conflicts.
\textbf{Stage II: Geospatial Supervised Fine-Tuning.}
We unfreeze the full architecture for supervised fine-tuning on remote sensing instruction datasets. This stage strengthens the model's ability to integrate visual evidence with linguistic instructions for geospatial reasoning.

\section{Experiments}
\label{sec:experiments}

In this section, we evaluate SkyNative on region-level understanding, large-format reasoning, robustness, and hallucination. Section~\ref{sec:exp_setup} introduces the benchmarks, evaluation metrics, baselines, and implementation details. Section~\ref{subsec:main} presents the main comparisons on standard and large-format remote sensing benchmarks, while Section~\ref{sec:robust_hallu} evaluates model robustness and hallucination. Finally, Section~\ref{subsec:ablation} examines the training strategy and architectural design, Section~\ref{subsec:visual_token_analysis} analyzes visual-token utilization, and Section~\ref{subsec:qualitative_results} presents qualitative examples.

\subsection{Experiment Setup}
\label{sec:exp_setup}

\textbf{Benchmarks and Evaluation Metrics.}
We evaluate SkyNative on a diverse suite of remote sensing benchmarks covering region-level understanding, large-format reasoning, and robustness-oriented analysis.
\textbf{(1) Region-Level Understanding.}
We use AID~\cite{xia2017aid} and RS19~\cite{xia2010structural} for scene classification, RSVQA-HR~\cite{lobry2020rsvqa} and VRSBench~\cite{li2024vrsbench} for remote sensing VQA, and DOTA-val~\cite{xia2018dota}, HRRSD~\cite{zhang2019hierarchical}, VHR~\cite{cheng2014multi}, and VisDrone~\cite{cao2021visdrone} for dense object counting. These benchmarks evaluate scene-level recognition, question answering, and fine-grained object-level perception on standard remote sensing image patches.
\textbf{(2) Large-Format Reasoning.}
We evaluate large-format remote sensing reasoning on XLRS~\cite{wang2025xlrs}, MME-RW-RS~\cite{zhang2025mmerealworld}, LRS-VQA~\cite{luo2025large}, and RSHR~\cite{dang2025benchmarkultrahighresolutionremotesensing}.
\textbf{(3) Robustness and Hallucination Analysis.}
We further include CHOICE~\cite{an2025choice}, OmniEarth~\cite{fu2026omniearth}, and HnstD~\cite{pang2025vhm} for controlled consistency, robustness, and hallucination-oriented analysis.

\textbf{Evaluation Metrics.}
To rigorously assess the performance of SkyNative, we utilize a comprehensive set of metrics covering perception, reasoning, and system efficiency.

(1) \textit{Accuracy (Acc.)}: Used for scene classification and VQA tasks, defined as the ratio of correctly predicted instances to the total number of samples.

(2) \textit{Mean Absolute Error (MAE)}: Employed for object counting tasks to measure the average absolute difference between the predicted count and the ground truth.

(3) \textit{Multi-Turn Exact Match (MTEM@1)}: To evaluate the consistency of multi-turn dialogues, we adopt the strict dialog-level ``all-correct'' metric:
For each dialogue $d$, let $c_{d,i}\in\{0,1\}$ indicate whether the prediction at turn $i$ exactly matches the reference answer. MTEM@1 is defined as
\[
\mathrm{MTEM@1}
=
\frac{100}{|\mathcal{D}|}
\sum_{d\in\mathcal{D}}
\mathbb{I}
\left[
\min_{1\leq i\leq T_d} c_{d,i}=1
\right],
\]
where a dialogue is considered correct only when all its valid turns are answered correctly.

\textbf{Baseline Models.}
We compare SkyNative with a diverse set of vision-language models across general and remote sensing domains.
\textbf{(1) General-Domain VLMs.}
We include closed-source models, namely Gemini-2.0~\cite{comanici2025gemini25pushingfrontier} and Claude-4~\cite{anthropic2025claude4}, as well as open-source models such as Qwen2.5-VL~\cite{bai2025qwen25vltechnicalreport}, DeepSeek-VL2~\cite{wu2024deepseekvl2}, MiniGPT-v2~\cite{chen2023minigptv2largelanguagemodel}, and VoCoT~\cite{li2025vocot}.
\textbf{(2) Native VLMs.}
We compare with general-domain native or encoder-free models, including EVEv2~\cite{diao2025evev2}, BREEN~\cite{li2026breen}, and Emu3~\cite{wang2026multimodal}.
\textbf{(3) Remote Sensing VLMs.}
We include representative remote sensing models, including GeoChat~\cite{kuckreja2024geochat}, VHM~\cite{pang2025vhm}, SkySenseGPT~\cite{luo2024skysensegptfinegrainedinstructiontuning}, and EarthDial~\cite{soni2025earthdial}.
\textbf{(4) Large-Format RS-VLMs.}
For large-format remote sensing imagery, we further compare with GeoLLaVA-8K~\cite{wang2025geollava}, Coarse-to-fine~\cite{luo2025large}, and ZoomEarth~\cite{liu2025zoomearth}.

\textbf{Implementation Details.}
We train SkyNative on two H200 GPUs, initialized from the EVE-7B-HD-v2.0 checkpoint. Training consists of two stages. The first stage uses 686K samples from RSVQA-LR~\cite{lobry2020rsvqa}, FloodNet~\cite{rahnemoonfar2020floodnethighresolutionaerial}, and GeoChat-instruction~\cite{kuckreja2024geochat}, while the second stage uses 380K samples from GeoCoT (w/o CoT)~\cite{liu2026faithfulreasoningremotesensing}. Each stage is trained for one epoch. We adopt DeepSpeed ZeRO-3, BF16 mixed precision, and AdamW with a cosine learning-rate scheduler and a warmup ratio of 0.03. For both stages, the maximum learning rate is $5\times10^{-6}$, the maximum sequence length is 4096, and the total batch size is 192. For visual processing, the any-ratio strategy preserves the aspect ratio and resizes each image to fit the sequence-length limit before converting it into non-overlapping $16\times16$ patch tokens with a stride of 16. The benchmark resolutions refer to the original images rather than the resized model inputs.

\begin{table*}[!htbp]
    \centering
    \caption{Quantitative evaluation of foundational perception on patch-level benchmarks.}
    \label{tab:patch_performance}
    \resizebox{\textwidth}{!}{%
    \begin{tabular}{l c c c c  c c  c c  c c  c c}
    \toprule
    \multirow{2}{*}{\textbf{Model}} & \textbf{AID} & \textbf{RS19} & \textbf{RSVQA-HR} & \textbf{VRSBench} & \multicolumn{2}{c}{\textbf{DOTA-val}} & \multicolumn{2}{c}{\textbf{HRRSD}} & \multicolumn{2}{c}{\textbf{VHR}} & \multicolumn{2}{c}{\textbf{VisDrone}} \\
    \cmidrule(lr){2-2} \cmidrule(lr){3-3} \cmidrule(lr){4-4} \cmidrule(lr){5-5}
    \cmidrule(lr){6-7} \cmidrule(lr){8-9} \cmidrule(lr){10-11} \cmidrule(lr){12-13}
    & \textit{Acc.} & \textit{Acc.} & \textit{Acc.} & \textit{Acc.} & \textit{Acc.} & \textit{MAE} & \textit{Acc.} & \textit{MAE} & \textit{Acc.} & \textit{MAE} & \textit{Acc.} & \textit{MAE} \\
    \midrule
    Gemini-2.0  & 76.00 & 90.00 & 17.00 & \underline{56.75} & 29.36 & 14.93 & 54.65 & 1.91 & \textbf{64.00} & 1.56 & 17.00 & \textbf{5.58} \\
    Claude-4 & 60.33 & 76.32 & 25.00 & 54.38 & 25.83 & 9.74 & 50.57 & 2.10 & 52.00 & 1.68 & 8.50 & 10.85 \\
    EVEv2 & 64.17 & 73.42 & 34.25 & 53.38 & 22.96 & 10.37 & 48.75 & 1.22 & 49.00 & 1.53 & 10.50 & 8.56 \\
    Qwen2.5-VL & 68.17 & 84.74 & 22.75 & 53.50 & \underline{33.77} & 9.73 & 57.82 & \underline{0.85} & 58.00 & \underline{1.17} & 13.50 & 7.15 \\
    MiniGPT-v2 & 22.50 & 28.42 & 23.50 & 34.88 & 10.82 & 56.91 & 19.50 & 13.60 & 21.00 & 4.66 & 10.00 & 9.37 \\
    VoCoT  & 46.67 & 76.84 & \underline{40.25} & 44.63 & 20.97 & 9.06 & 41.72 & 1.38 & 41.00 & 2.36 & 6.50 & 9.89 \\
    VHM & \underline{79.00} & \underline{91.84} & 36.50 & 54.00 & 32.67 & 9.26 & 46.71 & 1.05 & 48.50 & 1.26 & 6.50 & 8.01 \\
    SkySenseGPT & 75.50 & \textbf{93.16} & 36.25 & 51.88 & 33.11 & \underline{7.20} & 58.73 & 1.05 & 49.50 & 1.89 & \underline{18.50} & \underline{6.50} \\
    EarthDial & 67.33 & 73.42 & \textbf{42.00} & 42.00 & 32.23 & 8.42 & \underline{61.45} & 0.87 & 52.50 & 1.33 & 18.00 & 8.41 \\
    \midrule
     \rowcolor{blue!10}
     \textbf{SkyNative (Ours)} & \textbf{84.83} & 90.00 & \underline{40.25} & \textbf{69.38} & \textbf{35.98} & \textbf{5.87} & \textbf{68.93} & \textbf{0.61} & \underline{60.50} & \textbf{0.99} & \textbf{19.00} & 7.20 \\
    \bottomrule
    \end{tabular}
    }
\end{table*}

\subsection{Main Results and Analysis}
\label{subsec:main}

In this section, we comprehensively evaluate SkyNative. We first assess its foundational perception and reasoning capabilities on patch-level imagery, followed by a rigorous evaluation of its fine-grained reasoning performance on large-format remote sensing benchmarks.

\begin{table}[t]
    \centering
    \caption{Comparison on large-format remote sensing benchmarks.}
    \label{tab:uhr_performance}
    \scriptsize
    \setlength{\tabcolsep}{2.6pt}
    \renewcommand{\arraystretch}{1}
    \resizebox{\columnwidth}{!}{%
    \begin{tabular}{lcccccc}
    \toprule
    \multirow{2}{*}{\textbf{Method}}
    & \multirow{2}{*}{\textbf{XLRS}}
    & \multirow{2}{*}{\textbf{MME}}
    & \multirow{2}{*}{\textbf{LRS}}
    & \multicolumn{3}{c}{\textbf{RSHR}} \\
    \cmidrule(lr){5-7}
    & & & & \textit{Perc.} & \textit{Reas.} & \textit{MTEM} \\
    \midrule
    DeepSeek-VL2 & 38.31 & 27.74 & \underline{26.28} & \textbf{34.74} & 33.64 & 22.81 \\
    EVEv2 & \underline{38.41} & 40.07 & 23.04 & 28.34 & 42.80 & 23.68 \\
    BREEN & 37.02 & 24.67 & 23.22 & 30.22 & 30.91 &  9.65 \\
    Emu3 & 32.53 & 39.70 & 25.06 & 31.24 & 30.00 & 11.40 \\
    GeoChat & 26.56 & 28.12 & 13.72 & 28.39 & 35.45 & 13.16 \\
    VHM & 28.25 & 22.95 & 16.49 & 27.15 & 24.55 & 12.28 \\
    GeoLLaVA-8K & 35.89 & 20.63 & 21.87 & 28.91 & \underline{46.82} & \underline{26.32} \\
    Coarse-to-fine & 35.65 & 41.36 & 25.33 & 26.79 & 26.82 &  2.63 \\
    ZoomEarth & 30.65 & \underline{43.15} & 19.89 & 31.80 & 43.64 & 21.05 \\
    \midrule
    \rowcolor{blue!10}
    \textbf{SkyNative (Ours)} & \textbf{39.62} & \textbf{43.63} & \textbf{31.32} & \underline{32.02} & \textbf{47.40} & \textbf{28.07} \\
    \bottomrule
    \end{tabular}
    }
\end{table}

\textbf{Region-Level Remote Sensing Understanding.}
We first evaluate SkyNative on eight standard patch-level benchmarks, covering scene classification, visual question answering, and dense object counting. These benchmarks assess region-level remote sensing understanding from complementary perspectives, including scene-level semantics, language-conditioned visual reasoning, and fine-grained object-level perception.

\textbf{(1) Scene Classification.}
As shown in Table~\ref{tab:patch_performance}, SkyNative achieves the highest accuracy among the compared methods on AID (84.83\%) and competitive performance on RS19 (90.00\%). These results indicate that SkyNative can effectively capture scene-level semantics in remote sensing image regions.
\textbf{(2) Visual Question Answering.}
SkyNative achieves the highest score among the compared methods on VRSBench (69.38\%) and the second-best performance on RSVQA-HR (40.25\%). These results show that SkyNative maintains competitive performance in language-conditioned region-level understanding.
\textbf{(3) Dense Object Counting.}
SkyNative achieves the best overall results among the compared methods on DOTA-val (35.98\% Acc., 5.87 MAE) and HRRSD (68.93\% Acc., 0.61 MAE). It also performs competitively on VHR (60.50\% Acc., 0.99 MAE) and VisDrone (19.00\% Acc., 7.20 MAE). These results suggest that SkyNative is effective for fine-grained object-level perception, particularly in scenes containing small targets and dense object distributions.
Taken together, the results demonstrate the strong overall capability of SkyNative across different levels of region-level remote sensing understanding.

\textbf{Large-Format Remote Sensing Reasoning.}
Real-world remote sensing interpretation often requires models to identify local visual details while reasoning over broader spatial contexts. To evaluate this capability, we conduct experiments on four large-format remote sensing benchmarks: XLRS-Bench, MME-RW-RS, LRS-VQA, and RSHR-Bench. As shown in Table~\ref{tab:large_format_results}, SkyNative achieves scores of 39.62\% on XLRS-Bench, 43.63\% on MME-RW-RS, and 31.32\% on LRS-VQA. On RSHR-Bench, it obtains a reasoning accuracy of 47.40\% and an MTEM@1 score of 28.07\%, both of which are the highest among the compared methods.

These results demonstrate the strong overall capability of SkyNative in processing large-format remote sensing imagery. Its consistent performance across perception, reasoning, and multi-turn dialogue tasks indicates that the model can capture localized visual cues while maintaining broader spatial context. This behavior is consistent with the design goal of processing native visual tokens and textual information within a unified autoregressive backbone, supporting the application of native visual-token modeling to complex large-format remote sensing understanding.

\subsection{Robustness and Hallucination Analysis}
\label{sec:robust_hallu}

We further evaluate SkyNative under challenging remote sensing conditions, including degraded inputs, semantic consistency checks, and hallucination-oriented queries. We use OmniEarth-Robustness, which comprises Image Condition Assessment (ICA), Degraded-condition VQA (DVQA), Hallucination Detection (HD), and Semantic Consistency (SEC), as well as HnstD, which evaluates hallucination behavior through presence, position, and color questions.

As shown in Table~\ref{tab:performance_comparison}, SkyNative achieves the highest scores among the compared methods on all four OmniEarth-Robustness subtasks, with particularly strong performance on ICA and DVQA. On HnstD, SkyNative obtains a competitive average score of 62.13 and performs favorably on position and color questions, reaching 45.19 and 88.62, respectively. VHM achieves the highest overall HnstD score, which is consistent with its honesty-oriented design~\cite{pang2025vhm}. Taken together, these results show that SkyNative remains effective under degraded visual inputs and semantic consistency tests, while the HnstD results also indicate that reducing hallucinations across different question types remains a challenge.

\begin{table}[t]
    \centering
    \caption{Robustness and hallucination evaluation on OmniEarth-Robustness and HnstD.}
    \label{tab:performance_comparison}
    \renewcommand{\arraystretch}{1.05}

    \begin{tabular*}{\columnwidth}{@{\extracolsep{\fill}}lcccc@{}}
        \toprule
        \multicolumn{5}{c}{\textbf{OmniEarth-Robustness}} \\
        \midrule
        \textbf{Method} & \textbf{ICA} & \textbf{DVQA} & \textbf{HD} & \textbf{SEC} \\
        \midrule
        VHM          & 27.10 & 37.80 & 26.40 & 26.10 \\
        GeoChat      & 25.80 & 25.30 & 31.00 & 25.90 \\
        GeoLLaVA-8K  & 24.80 & 23.60 & 19.90 & \underline{27.60} \\
        SkySenseGPT  & 24.90 & 28.40 & \underline{34.80} & 25.80 \\
        \textbf{SkyNative (Ours)} & \textbf{50.00} & \textbf{63.43} & \textbf{38.00} & \textbf{28.67} \\
        \midrule[0.8pt]
        \multicolumn{5}{c}{\textbf{HnstD}} \\
        \midrule
        \textbf{Method} & \textbf{Presence} & \textbf{Position} & \textbf{Color} & \textbf{Avg.} \\
        \midrule
        VHM$^\ast$
            & \textcolor{gray!60}{87.07}
            & \textcolor{gray!60}{74.25}
            & \textcolor{gray!60}{91.50}
            & \textcolor{gray!60}{84.27} \\
        GeoChat      & 53.74 & 23.19 & 50.25 & 42.39 \\
        GeoLLaVA-8K  & \textbf{71.84} & 37.00 & 50.25 & 53.03 \\
        SkySenseGPT  & \underline{58.33} & 13.92 & 71.50 & 47.92 \\
        \textbf{SkyNative (Ours)} & 52.59 & \textbf{45.19} & \textbf{88.62} & \textbf{62.13} \\
        \bottomrule
    \end{tabular*}

    \begin{minipage}{\columnwidth}
        $^\ast$ denotes that VHM uses the honesty-oriented instruction data introduced with HnstD~\cite{pang2025vhm}.
    \end{minipage}
\end{table}

\begin{table}[t]
    \centering
    \small
    \caption{Ablation results of the two-stage training strategy.}
    \label{tab:stages_comparison_left}

    \setlength{\tabcolsep}{4pt}
    \renewcommand{\arraystretch}{1.05}

    \resizebox{\columnwidth}{!}{%
        \begin{tabular}{l ccccc}
            \toprule
            \textbf{Setting}
            & \textbf{LRS}
            & \textbf{RS19}
            & \textbf{VRSBench}
            & \multicolumn{2}{c}{\textbf{HRRSD}} \\
            \cmidrule(lr){2-2}
            \cmidrule(lr){3-3}
            \cmidrule(lr){4-4}
            \cmidrule(lr){5-6}

            & \textit{Acc.}
            & \textit{Acc.}
            & \textit{Acc.}
            & \textit{Acc.}
            & \textit{MAE} \\
            \midrule

            Baseline
            & 23.04 & 73.42 & 53.38 & 48.75 & 1.22 \\

            + Stage I
            & 23.21 & 74.74 & 55.50 & 63.04 & 0.90 \\

            + Stage II
            & \textbf{31.32}
            & \textbf{90.00}
            & \textbf{69.38}
            & \textbf{68.93}
            & \textbf{0.61} \\

            \bottomrule
        \end{tabular}%
    }
    
\end{table}

\subsection{Ablation Study}
\label{subsec:ablation}

\textbf{Training Strategy Ablation.}
To examine whether the proposed two-stage training process contributes to the final performance, we compare the baseline model with models trained after Stage I and Stage II. As shown in Table~\ref{tab:stages_comparison_left}, Stage I improves the results on most benchmarks, with a notable gain on HRRSD counting, where the accuracy increases from 48.75\% to 63.04\% and the MAE decreases from 1.22 to 0.90. After Stage II, the model achieves further improvements across all evaluated benchmarks, including LRS-VQA, RS19, VRSBench, and HRRSD.

These results indicate that the two-stage training process provides consistent performance gains over the baseline. In particular, Stage I mainly improves the model's adaptation to remote sensing data, while Stage II further enhances its performance on downstream remote sensing vision-language tasks. This suggests that both stages are useful for training SkyNative, and their combination leads to the best overall performance.

\begin{figure}[t]
    \centering
    \includegraphics[width=\columnwidth]{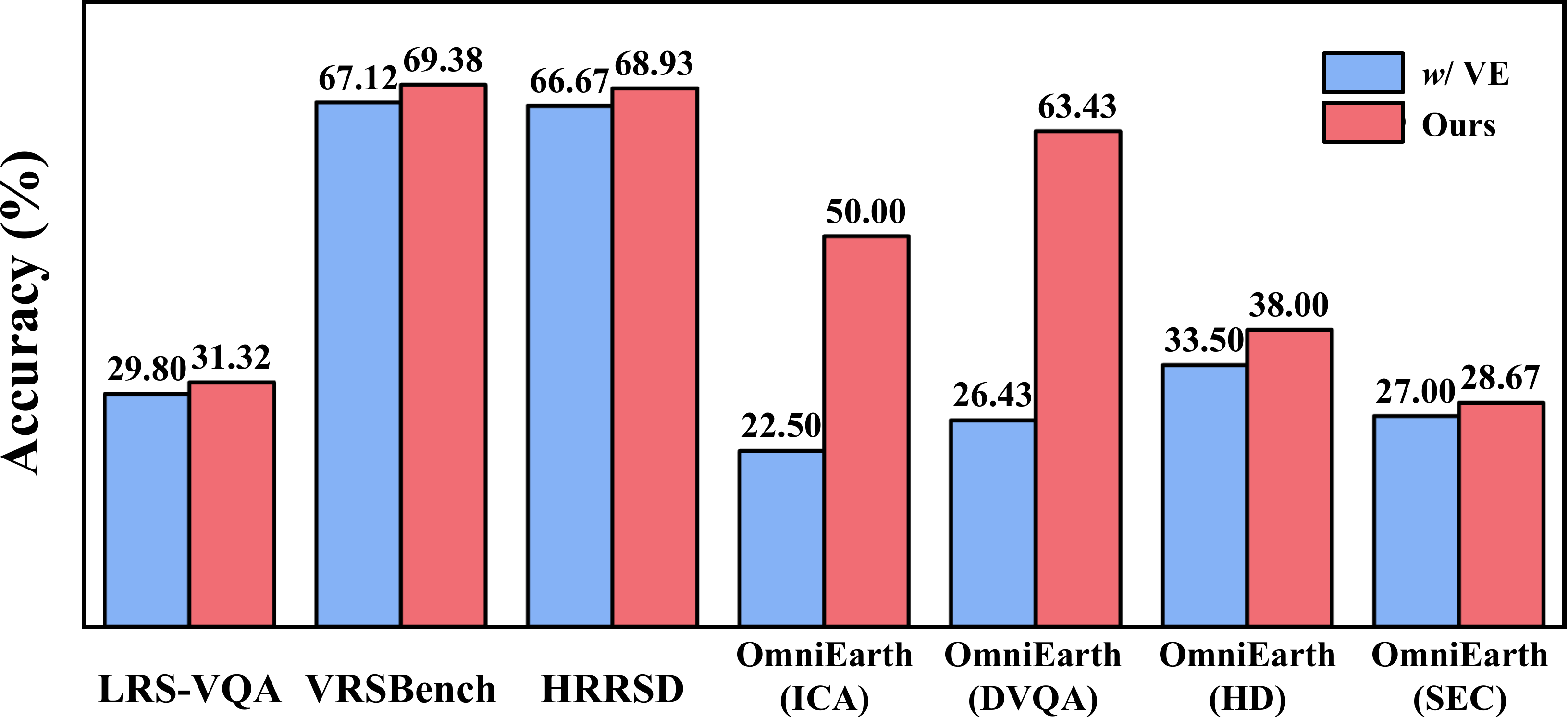}
    \caption{
    Architecture ablation under the same data and training settings.
    }
    \label{fig:architecture_ablation}
\end{figure}

\textbf{Model Architecture Ablation.} To analyze the effect of the proposed native VLM architecture, we compare SkyNative with a controlled CLIP-MLP variant. The CLIP-MLP variant follows a standard encoder-based visual pathway, where image features are extracted by a CLIP-ViT-L/336 vision encoder and mapped into the language embedding space through an MLP projector. SkyNative adopts a native VLM architecture. Both models are trained with identical training datasets, training strategy, optimization settings, and evaluation protocol.
Fig.~\ref{fig:architecture_ablation} reports the results on representative benchmarks, including LRS-VQA, VRSBench, HRRSD, and OmniEarth-Robustness. SkyNative achieves consistent improvements over the CLIP-MLP variant across these benchmarks. The improvements are moderate on standard understanding tasks, such as LRS-VQA, VRSBench, and HRRSD, and are more pronounced on robustness-oriented OmniEarth subtasks. These results suggest that the native VLM architecture of SkyNative is well adapted to remote sensing vision-language tasks under identical data and training settings.

\subsection{Visual-Token Utilization Analysis}
\label{subsec:visual_token_analysis}

\begin{figure}[t]
    \centering
    \includegraphics[width=\linewidth]{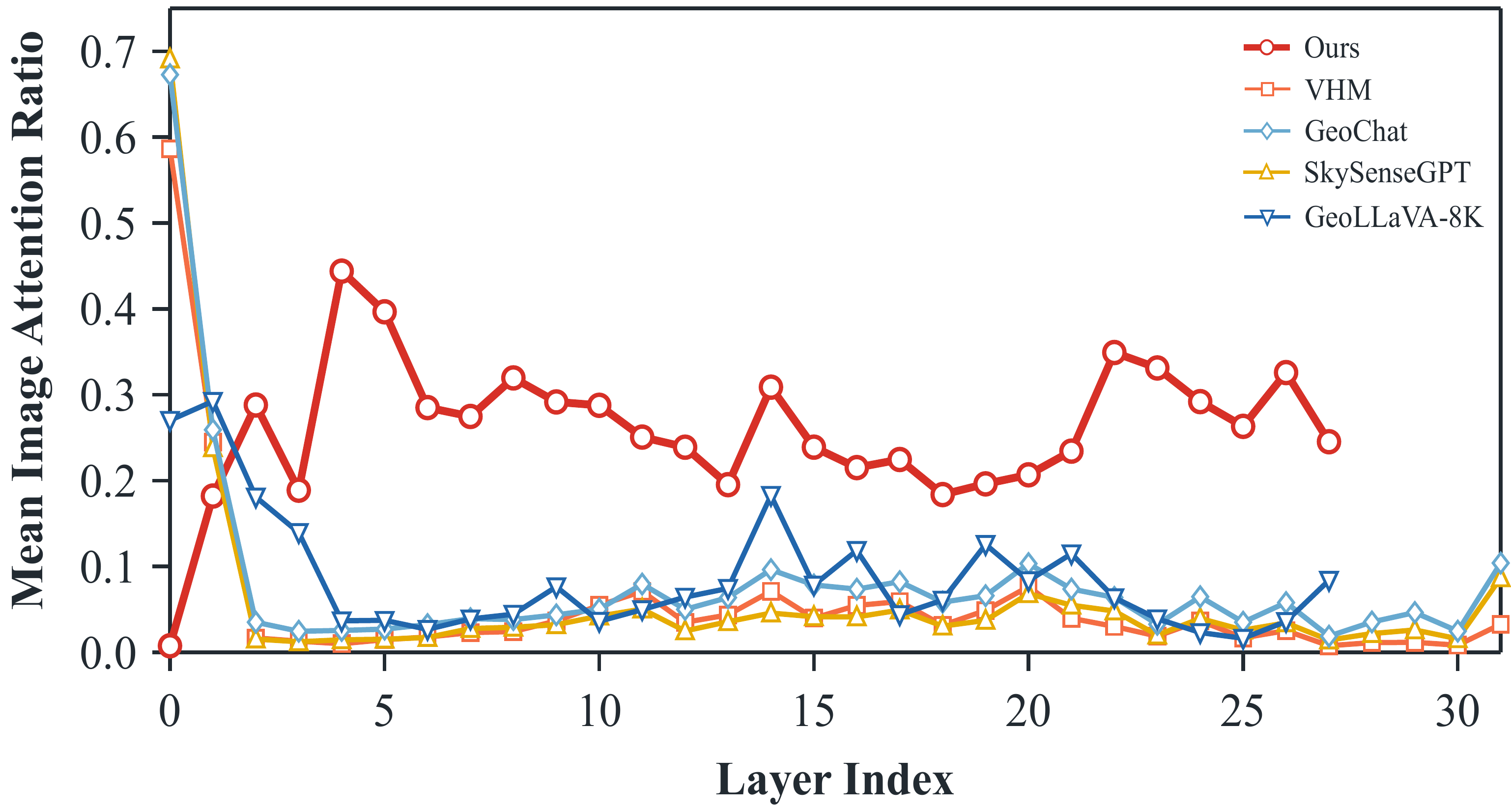}
    \caption{Layer-wise mean attention ratio from text-query tokens to image-key tokens.}
    \label{fig:attention_ratio}
\end{figure}

\begin{figure}[t]
    \centering
    \includegraphics[width=\linewidth]{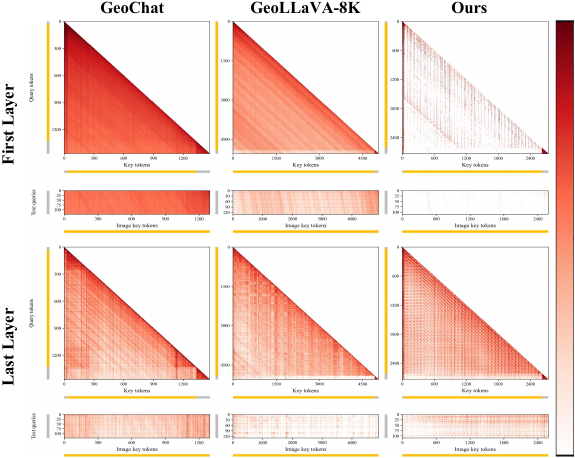}
    \caption{First- and last-layer causal attention maps under the same input. Yellow and gray denote image and text/special tokens, respectively, while the lower panels show text-to-image attention. Darker red indicates higher normalized attention.}
    \label{fig:attention_maps}
\end{figure}

We explore visual-information flow through its layer-wise allocation and token-level distribution. As shown in Fig.~\ref{fig:attention_ratio}, the encoder-based models allocate considerable attention to image tokens in shallow layers, followed by a rapid decrease through the language backbone. SkyNative exhibits a different trajectory, allocating a relatively high proportion of attention to visual tokens across most intermediate and deep layers. This pattern shows that visual tokens continue to receive substantial attention during the later stages of autoregressive inference. Since the evaluated models differ in visual tokenization and sequence length, we focus primarily on their layer-wise trends rather than treating the absolute ratios as a strictly controlled architectural comparison.

Figure~\ref{fig:attention_maps} further illustrates the token-level distribution of visual attention. SkyNative exhibits sparse but structured text-to-image interactions, particularly in the last layer, despite its relatively high overall visual attention ratio. This indicates that its attention is concentrated on a subset of visual tokens rather than uniformly distributed across the image. Such selective allocation is consistent with the characteristics of remote sensing imagery, where informative local objects and spatial relations must often be distinguished from extensive background regions. Together with the preceding benchmark results, these observations provide a complementary view of how native visual tokens are integrated into SkyNative's inference process.

\subsection{Qualitative Results}
\label{subsec:qualitative_results}

\begin{figure}[t]
    \centering\includegraphics[width=\linewidth]{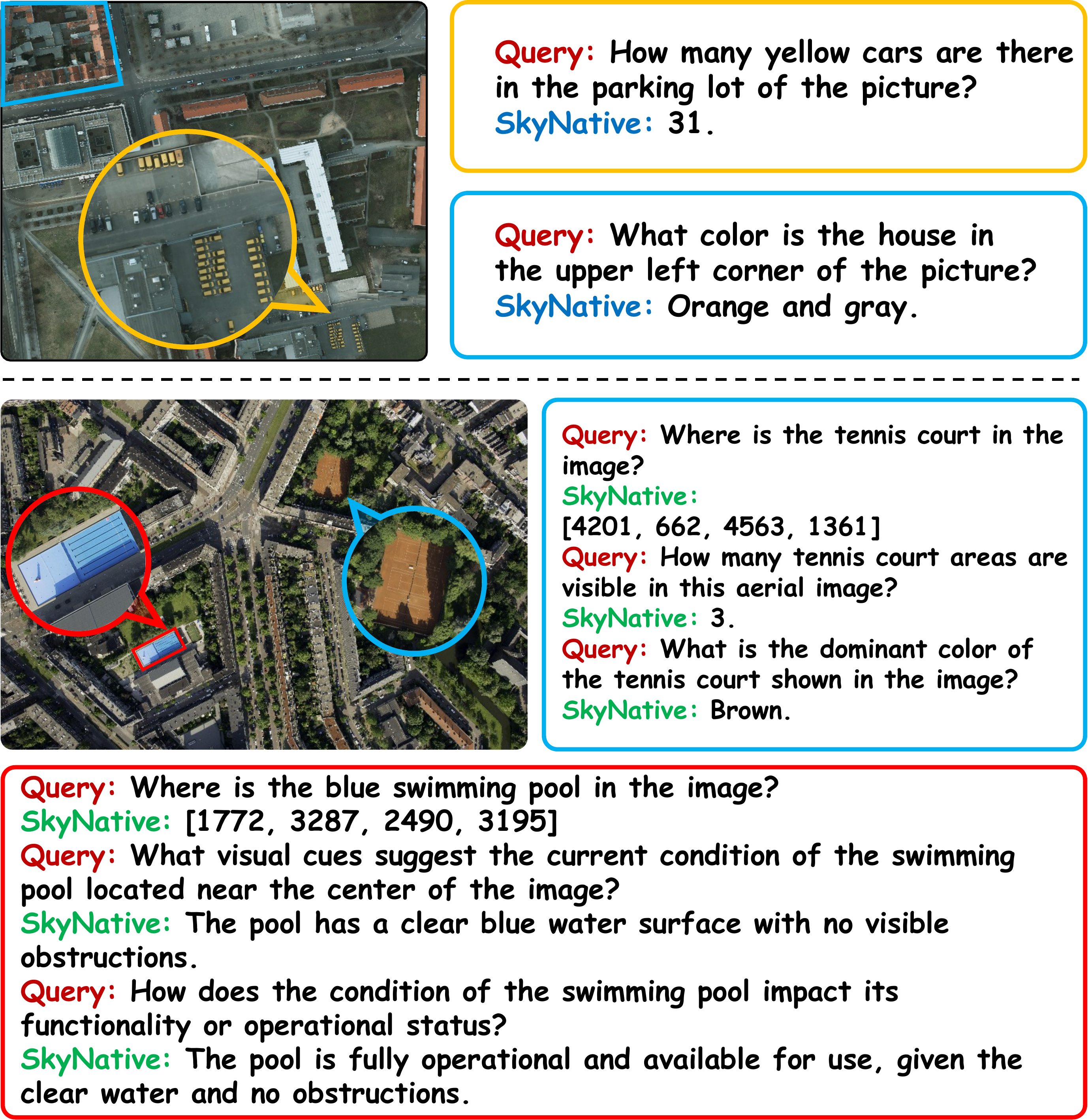} 
    \caption{Qualitative results of SkyNative on complex remote sensing imagery.}
    \label{fig:visual_results}
\end{figure}

To provide an intuitive analysis of SkyNative, we present qualitative examples in Fig.~\ref{fig:visual_results}. The examples cover several representative remote sensing vision-language tasks, including dense object counting, local attribute recognition, object localization, and multi-turn spatial reasoning. SkyNative accurately counts small yellow cars in a dense parking area and recognizes the color attributes of buildings in local regions, showing its ability to handle fine-grained visual details in complex remote sensing scenes.
SkyNative also shows consistent performance on localization and region-level reasoning. As shown in the swimming pool and tennis court examples, the model can identify target regions with coordinate outputs, answer follow-up questions about object quantity and dominant color, and reason about object conditions based on local visual context. These qualitative results are consistent with the quantitative results, indicating that SkyNative can handle diverse remote sensing understanding tasks involving small objects, local attributes, spatial localization, and multi-turn reasoning.

\section{Conclusion}
\label{sec:conclusion}
In this work, we explore native multimodal modeling for remote sensing vision--language tasks and present SkyNative. Instead of separating visual representation from language reasoning through a pretrained vision encoder and projector, SkyNative converts images into native visual tokens, prepends them to text tokens, and processes the resulting sequence within a unified autoregressive Transformer. A modality-aware decoupled design accommodates the different characteristics of vision and language within the shared architecture. The combined quantitative and attention analyses show that this design allows textual reasoning to access visual information across multiple layers of the language backbone, supporting fine-grained perception, large-format understanding, and reliable visual grounding. These findings highlight the value of providing textual reasoning with direct access to fine-grained visual context and support native multimodal modeling as a promising architectural direction for remote sensing vision--language models.

\bigskip

\bibliography{aaai2027}

\end{document}